\documentclass[manuscript,screen]{acmart}

\AtBeginDocument{%
  \providecommand\BibTeX{{%
    \normalfont B\kern-0.5em{\scshape i\kern-0.25em b}\kern-0.8em\TeX}}}

\copyrightyear{2021}
\acmYear{2021}
\setcopyright{acmlicensed}
\acmConference[RecSys '21]{Fifteenth ACM Conference on Recommender Systems}{September 27-October 1, 2021}{Amsterdam, Netherlands}
\acmBooktitle{Fifteenth ACM Conference on Recommender Systems (RecSys '21), September 27-October 1, 2021, Amsterdam, Netherlands}
\acmPrice{15.00}
\acmDOI{10.1145/3460231.3474252}
\acmISBN{978-1-4503-8458-2/21/09}

\begin{document}

\title{Cold Start Similar Artists Ranking with Gravity-Inspired~Graph~Autoencoders}

\author{Guillaume Salha-Galvan}
\authornote{Contact author: \href{research@deezer.com}{research@deezer.com}}
\affiliation{
  \institution{Deezer Research \& LIX, \'{E}cole Polytechnique}
  \city{}
  \country{France}
}

\author{Romain Hennequin}
\affiliation{
  \institution{Deezer Research}
    \city{}
  \country{France}
}

\author{Benjamin Chapus}
\affiliation{
  \institution{Deezer Research}
    \city{}
  \country{France}
}

\author{Viet-Anh Tran}
\affiliation{
  \institution{Deezer Research}
    \city{}
  \country{France}
}

\author{Michalis Vazirgiannis}
\affiliation{
  \institution{LIX, \'{E}cole Polytechnique}
    \city{}
  \country{France}
}

\renewcommand{\shortauthors}{G. Salha-Galvan et al.}

\newcommand{\up}[1]{\textsuperscript{#1}}

\begin{abstract}
On an artist's profile page, music streaming services frequently recommend a ranked list of \textit{"similar artists"} that fans also liked.
However, implementing such a feature is challenging for new artists, for which usage data on the service (e.g. streams or likes) is not yet available.
In this paper, we model this \textit{cold start similar artists ranking} problem as a link prediction task in a directed and attributed graph, connecting artists to their top-$k$ most similar neighbors and incorporating side musical information.
Then, we leverage a graph autoencoder architecture to learn node embedding representations from this graph, and to automatically rank the top-$k$ most similar neighbors of new artists using a gravity-inspired mechanism.
We empirically show the flexibility and the effectiveness of our framework, by addressing a real-world cold start similar artists ranking problem on a global music streaming service. Along with this paper, we also publicly release our source code as well as the industrial graph data from our experiments.
\end{abstract}

\ccsdesc[300]{Information systems~Recommender systems}
\ccsdesc[300]{Computing methodologies~Learning latent representations}
\ccsdesc[300]{Computing methodologies~Ranking}
\ccsdesc[300]{Mathematics of computing~Graph algorithms}

\keywords{Music Recommendation, Music Streaming Services, Cold Start, Similar Music Artists, Ranking, Directed Graphs, Autoencoders, Variational Autoencoders, Graph Representation Learning, Node Embedding, Link Prediction}

\maketitle

\section{Introduction}

Music streaming services heavily rely on recommender systems to help users discover and enjoy new musical content within large catalogs of millions of songs, artists and albums, with the general aim of improving their experience and engagement \cite{briand2021semi,schedl2018current,bendada2020carousel,mehrotra2019jointly}. 
In particular, these services frequently recommend, on an artist's profile page, a ranked list of related artists that fans also listened to or liked \cite{donker2019networking,spotify2019fansalsolike,kjus2016musical}. 
Referred to as \textit{"Fans Also Like"} on Spotify and Soundcloud and as \textit{"Related"} or \textit{"Similar Artists"} on Amazon Music, Apple Music and Deezer, such a feature typically leverages \textit{learning to rank} models \cite{karatzoglou2013learning,schedl2018current,rafailidis2017learning}. It retrieves the most relevant artists according to similarity measures usually computed from usage data, e.g. from the proportion of shared listeners across artists \cite{donker2019networking,spotify2019fansalsolike}, or from more complex \textit{collaborative filtering} models \cite{jain2020survey,koren2015advances,schedl2018current} that predict similarities from the known preferences of an artist’s listeners.
It has recently been described as \textit{"one of the easiest ways"} to let \textit{"users discover new music"} by Spotify \cite{spotify2019fansalsolike}.

However, filling up such ranked lists is especially challenging for new artists.
Indeed, while music streaming services might have access to some general descriptive information on these artists, listening data will however not be available upon their first release.
This prevents computing the aforementioned usage-based similarity measures.
As a consequence of this problem, which we refer to as \textit{cold start similar artists ranking}, music streaming services usually do not propose any \textit{"Fans Also Like"} section for these artists, until (and if ever) a sufficiently large number of usage interactions, e.g. listening sessions, has been reached. 
Besides new artists, this usage-based approach also excludes from recommendations a potentially large part of the existing catalog with too few listening data, which raises fairness concerns~\cite{corbett2018measure}. 
Furthermore, while we will focus on music streaming applications, this problem encompasses the more general \textit{cold start similar items ranking} issue, which is also crucial for media recommending other items~such~as~videos~\cite{youtube}.

In this paper, we address this problem by exploiting the fact that, as detailed in Section \ref{s3}, such \textit{"Fans Also Like"} features can naturally be summarized as a directed and attributed \textit{graph}, that connects each item \textit{node}, e.g. each artist, to their most similar neighbors via directed \textit{links}. Such a graph also incorporates additional descriptive information on nodes and links from the graph, e.g. musical information on artists. In this direction, we model cold start similar items ranking as a \textit{directed link prediction} problem \cite{salha2019-2}, for new nodes gradually added into this graph.

Then, we solve this problem by leveraging recent advances in graph representation learning \cite{hamilton2017representation,wu2019comprehensive,hamilton2020graph}, and specifically directed \textit{graph autoencoders} \cite{kipfvgae,salha2019-2}. Our proposed framework permits retrieving similar neighbors of items from \textit{node embeddings} and from a \textit{gravity-inspired} decoder acting as a ranking mechanism. 
Our work is the first transposition and analysis of gravity-inspired graph autoencoders \cite{salha2019-2} on recommendation problems.
Backed by in-depth experiments on industrial data from the global music streaming service Deezer\footnote{\href{https://www.deezer.com}{https://www.deezer.com}}, we show the effectiveness of our approach at addressing a real-world cold start similar artists ranking problem, outperforming several popular baselines for cold start recommendation.
Last, we publicly release our code and the industrial data from our experiments. Besides making our results reproducible, we hope that such a release of real-world resources will benefit future~research~on~this~topic.

This paper is organized as follows. In Section \ref{s2}, we introduce the cold start similar items ranking problem more precisely and mention previous works on related topics. In Section \ref{s3}, we present our graph-based framework to address this problem. We report and discuss our experiments on Deezer data in Section \ref{s4}, and we conclude in Section \ref{s5}.

\section{Preliminaries}
\label{s2}

\subsection{Problem Formulation}
\label{s21}

Throughout this paper, we consider a catalog of $n$ recommendable items on an online service, such as music artists in our application. Each item $i$ is described by some side information summarized in an $f$-dimensional vector $x_i$; for artists, such a vector could for instance capture information related to their country of origin or to their music genres. These $n$~items are assumed to be \textit{"warm"}, meaning that the service considers that a sufficiently large number of interactions with users, e.g. likes or streams, has been reached for these items to ensure reliable usage data analyses.

From these usage data, the service learns an $n \times n$ \textit{similarity matrix} $S$, where the element $S_{ij} \in [0, 1]$ captures the similarity of item $j$ w.r.t. item $i$. Examples of some possible usage-based similarity scores\footnote{Details on the computation of similarities at Deezer are provided in Section~\ref{s41} - without loss of generality, as our framework is valid~for~any~$S_{ij} \in [0, 1]$.} $S_{ij}$ include the percentage of users interacting with item $i$ that also interacted with item $j$ (e.g. users listening to or liking both items \cite{spotify2019fansalsolike}), mutual information scores \cite{shakibian2017mutual}, or more complex measures derived from collaborative filtering \cite{jain2020survey,koren2015advances,schedl2018current}. Throughout this paper, we assume that similarity scores are fixed over time, which we later discuss.

Leveraging these scores, the service proposes a \textit{similar items} feature comparable to the \textit{"Fans Also Like"} described in the introduction. Specifically, along with the presentation of an item $i$ on the service, this feature recommends a ranked list of $k$ similar items to users. They correspond to the top-$k$ items $j$ such as $j \neq i$ and with highest similarity scores $S_{ij}$.

On a regular basis, \textit{"cold"} items will appear in the catalog. While the service might have access to descriptive side information on these items, \textit{no usage data} will be available upon their first online release. This hinders the computation of usage-based similarity scores, thus excluding these items from recommendations until they become warm - if ever. 
In this paper, we study the feasibility of effectively predicting their future similar items ranked lists, from the delivery of these items i.e. without any usage data. This would enable offering such an important feature quicker and on a larger part of the catalog. More precisely, we answer the following research question: using only 1) the known similarity scores between warm items, and 2) the available descriptive information, how, and to which extent, can we predict the future \textit{"Fans Also Like"} lists that would ultimately be computed~once cold items become warm?

\subsection{Related Work}
\label{s22}

While collaborative filtering methods effectively learn item proximities, e.g. via the factorization of user-item
interaction matrices \cite{van2013deep,koren2015advances}, these methods usually become unsuitable for cold items without any interaction data and thus~absent from these matrices \cite{van2013deep}. In such a setting, the simplest strategy for similar items ranking would consist in relying on \textit{popularity} metrics \cite{schedl2018current}, e.g. to recommend the most listened artists. In the presence of descriptive information on cold items, one could also recommend items with the closest descriptions \cite{abbasifard2014survey}. These heuristics are usually outperformed by \textit{hybrid} models, leveraging both item descriptions and collaborative filtering on warm items~\cite{van2013deep,hsieh2017collaborative,wang2018billion,he2016vbpr}.~They~consist~in:
\begin{itemize}
    \item learning a vector space representation (an \textit{embedding}) of warm items, where proximity~reflects~user~preferences;
    \item  then, projecting cold items into this embedding, typically by learning a model to map descriptive vectors of warm items to their embedding vectors, and then applying this mapping to cold items' descriptive vectors.
\end{itemize}

Albeit under various formulations, this strategy has been transposed to Matrix Factorization \cite{van2013deep,briand2021semi}, Collaborative Metric Learning \cite{hsieh2017collaborative,lee2018collaborative} and Bayesian Personalized Ranking \cite{he2016vbpr,barkan2019cb2cf}; in practice, a deep neural network often acts as the mapping model. The retrieved \textit{similar items} are then the closest ones in the embedding. Other deep learning approaches were also recently proposed for item cold start, with promising performances. DropoutNet \cite{volkovs2017dropoutnet} processes both usage and descriptive data, and is explicitly trained for cold start through a dropout \cite{srivastava2014dropout} simulation mechanism. MeLU \cite{lee2019melu} deploys a meta-learning paradigm to learn embeddings in the absence of usage data. CVAE \cite{li2017collaborative} leverages a Bayesian generative process to sample cold embedding vectors, via a collaborative~variational~autoencoder.

While they constitute relevant baselines, these models do not rely on \textit{graphs}, contrary to our work. Graph-based recommendation has recently grown at a fast pace (see the surveys of \cite{wang2021graph,wu2020graph}), including in industrial applications~\cite{wang2018billion,ying2018graph}. Existing research widely focuses on bipartite user-item graphs \cite{wang2021graph}. Notably, STAR-GCN \cite{zhang2019star} addresses cold start by reconstructing user-item links using stacked graph convolutional networks, extending ideas from \cite{berg2018graph,kipfvgae}. Instead, recent efforts \cite{qian1,qian2} emphasized the relevance of leveraging - as we will - graphs connecting items together, along with their attributes. In this direction, the work closest to ours might be the recent DEAL \cite{ijcai2020DEAL} who, thanks to an alignment mechanism, predicts links in such graphs for new nodes having only descriptive information.
We will also compare to DEAL; we nonetheless point out that their work focused on undirected graphs, and did not consider~ranking~settings.

\section{A Graph-Based Framework for Cold Start Similar Items Ranking}
\label{s3}

In this section, we present our proposed graph-based framework to address the cold start similar items ranking problem.

\subsection{Similar Items Ranking as a Directed Link Prediction Task}
\label{s31}

We argue that \textit{"Fans Also Like"} features can naturally be summarized as a graph structure with $n$ nodes and $n \times k$ edges. Nodes are warm recommendable items from the catalog, e.g. music artists with enough usage data according to the service's internal rules. Each item node points to its $k$ most similar neighbors via a link, i.e. an edge. This graph is:
\begin{itemize}
    \item \textit{directed}: edges have a direction, leading to asymmetric relationships. For instance,
    while most fans of a little known reggae band might listen to Bob Marley (Marley thus appearing among their similar artists), Bob Marley's fans will rarely listen to this band, which is unlikely to appear back among Bob Marley’s own similar artists. 
    \item \textit{weighted}: among the $k$ neighbors of node $i$, some items are more similar to $i$ than others (hence the ranking). We capture this aspect by equipping each directed edge $(i,j)$ from the graph with a \textit{weight} corresponding to the similarity score $S_{ij}$. More formally, we summarize our graph structure by the $n \times n$ \textit{adjacency matrix} $A$, where the element $A_{ij} = S_{ij}$ if $j$ is one of the $k$ most similar items w.r.t. $i$, and where $A_{ij} = 0$ otherwise\footnote{Alternatively, one could consider a \textit{dense} matrix where $A_{ij} = S_{ij}$ for all pairs $(i,j)$. However, besides acting as a data cleaning process on lowest scores, sparsifying $A$ speeds up computations for the encoder introduced thereafter, whose complexity evolves linearly w.r.t. the number of edges~(see~Section~\ref{s343}).}.
    \item \textit{attributed}: as explained in Section \ref{s21}, each item $i$ is also described by a vector $x_i \in \mathbb{R}^f$. In the following, we denote by $X$ the $n \times f$ matrix stacking up all descriptive vectors from the graph, i.e. the $i$-th row of $X$ is $x_i$.
\end{itemize}

Then, we model the release of a cold recommendable item in the catalog \textit{as the addition of a new node} in the graph, together with its side descriptive vector. As usage data and similarity scores are unavailable for this item, it is \textit{observed as isolated}, i.e. it does not point to $k$ other nodes. In our framework, we assume that these $k$ missing directed edges~-~and their weights~- are actually \textit{masked}. They point to the $k$ nodes with highest similarity scores, as would be identified by the service 
once it collects enough usage data to consider the item as warm, according to the service's criteria. These links and their scores, ultimately revealed, are treated as \textit{ground-truth} in the remainder of this work.

From this perspective, the cold start similar items ranking problem consists in a \textit{directed link prediction} task \cite{schall2014link,salha2019-2,lu2011link}. Specifically, we aim at predicting the locations and weights - i.e. estimated similarity scores - of these $k$ missing directed edges, and at comparing predictions with the actual ground-truth edges ultimately revealed, both in terms of:
\begin{itemize}
    \item \textit{prediction accuracy:} do we retrieve the correct locations of missing edges in the graph?
    \item \textit{ranking quality:} are the retrieved edges correctly ordered, in terms of similarity scores?
\end{itemize}

\subsection{From Similar Items Graphs to Directed Node Embeddings}
\label{s32}

Locating missing links in graphs has been the objective of significant research efforts from various fields \cite{schall2014link,salha2019-2,lu2011link,liben2007link}. While this problem has been historically addressed via the construction of hand-engineered node similarity metrics, such as the popular Adamic-Adar, Jaccard or Katz measures \cite{liben2007link}, significant improvements were recently achieved by methods directly \textit{learning} node representations summarizing the graph structure \cite{hamilton2017representation,wu2019comprehensive,kipfgcn,kipfvgae,hamilton2020graph}.
These methods represent each node $i$ as a vector $z_i \in \mathbb{R}^d$ (with $d \ll n$) in a \textit{node embedding space} where structural proximity should be preserved, and learned by using matrix factorization \cite{cao2015grarep},  random walks \cite{perozzi2014deepwalk}, or graph neural networks \cite{kipfgcn,wu2019comprehensive}. Then, they estimate the probability of a missing edge between two nodes by evaluating their proximity in~this~embedding~\cite{kipfvgae,hamilton2017representation}.

In this paper, we build upon these advances and thus learn node embeddings to tackle link prediction in our similar items graph. Specifically, we propose to leverage \textit{gravity-inspired graph (variational) autoencoders}, recently introduced in a previous work from Deezer \cite{salha2019-2}. As detailed in Section \ref{s33}, these models are doubly advantageous for our application:
\begin{itemize}
    \item Foremost, they can effectively \textit{process node attribute vectors} in addition to the graph, contrary to some popular alternatives such as DeepWalk  \cite{perozzi2014deepwalk} and standard Laplacian eigenmaps \cite{belkin2001laplacian}. This will help us add some cold nodes, isolated but equipped with some descriptions, into an existing warm node embedding.
    \item Simultaneously, they were specifically designed to address \textit{directed} link prediction from embeddings, contrary to the aforementioned alternatives or to standard graph autoencoders \cite{kipfvgae}. 
\end{itemize}
In the following, we first recall key concepts related to gravity-inspired graph (variational) autoencoders, in Section~\ref{s33}. Then, we transpose them to similar items ranking, in Section~\ref{s34}. To the best of our knowledge, our work constitutes the first analysis of these models on a recommendation problem and, more broadly, on~an~industrial~application.

\subsection{Gravity-Inspired Graph (Variational) Autoencoders}
\label{s33}

\subsubsection{Overview} 
\label{s331}

To predict the probability of a missing edge - and its weight - between two nodes $i$ and $j$ from some embedding vectors $z_i$ and $z_j$, most existing methods rely on symmetric measures, such as the Euclidean distance between these vectors $||z_i - z_j||_2$ \cite{hamilton2017representation} or their inner-product $z^T_i z_j$ \cite{kipfvgae}. However, due to this symmetry, predictions will be identical for both edge directions $i \rightarrow j$ and $j \rightarrow i$. This is undesirable for directed graphs, where $A_{ij} \neq A_{ji}$~in~general.

Salha et al. \cite{salha2019-2} addressed this problem by equipping $z_i$ vectors with \textit{masses}, and then taking inspiration from Newton’s second law of motion \cite{newton1687philosophiae}. Considering two objects $1$ and $2$ with positive masses $m_1$ and $m_2$, and separated by a distance $r>0$, the physical acceleration $a_{1 \rightarrow 2}$ (resp. $a_{2 \rightarrow 1}$) of $1$ towards $2$ (resp. $2$ towards $1$)~due~to~\textit{gravity} \cite{salha2019-2} is $a_{1 \rightarrow 2} = \frac{Gm_2}{r^2}$  (resp. $a_{2 \rightarrow 1}= \frac{Gm_1}{r^2}$), where $G$ denotes the gravitational constant \cite{cavendish1798xxi}.
One observes that $a_{1 \rightarrow 2} > a_{2 \rightarrow 1}$ when $m_2 > m_1$, i.e. that the acceleration of the less massive object towards the more massive one is higher.
Transposing these notions to node embedding vectors $z_i$ augmented with masses $m_i$ for each node $i$, Salha et al. \cite{salha2019-2} reconstruct \textit{asymmetric} links by using $a_{i \rightarrow j}$ (resp. $a_{j \rightarrow i}$) as an indicator of the likelihood that node $i$ is connected to $j$ (resp. $j$ to $i$) via a directed edge, setting $r = \|z_i - z_j\|_2$. Precisely, they use the logarithm of $a_{i \rightarrow j}$, limiting accelerations towards very central nodes \cite{salha2019-2}, and add a sigmoid activation $\sigma(x) = 1/(1 + e^{-x}) \in ]0,1[$. Therefore, for weighted graphs, the output is an estimation $\hat{A}_{ij}$ of some true weight $A_{ij} \in [0,1]$. For unweighted graphs, it corresponds to~the~probability~of~a~missing~edge:
\begin{equation}
\hat{A}_{ij} = \sigma(\log a_{i \rightarrow j}) = \sigma(\log \frac{G m_j}{\|z_i - z_j\|_2^2}) = \sigma(\underbrace{\log G m_j}_{\text{denoted } \tilde{m}_j} - \log \|z_i - z_j\|_2^2).
\label{equation2}
\end{equation}
\subsubsection{Gravity-Inspired Graph AE}
\label{s332}
To learn such masses and embedding vectors from an $n\times n$ adjacency matrix $A$, potentially completed with an $n\times f$ node attributes matrix $X$, Salha et al. \cite{salha2019-2} introduced \textit{gravity-inspired graph autoencoders}. They build upon the graph extensions of \textit{autoencoders} (AE) \cite{kipfvgae,tian2014learning,wang2016structural}, that recently appeared among the state-of-the-art approaches for (undirected) link prediction in numerous applications \cite{salha2020fastgae,salha2020simple,wang2016structural,kipfvgae,grover2019graphite,semiimplicit2019}. Graph~AE are a family of models aiming at \textit{encoding} nodes into an embedding space from which \textit{decoding} i.e. reconstructing the graph should ideally be possible, as, intuitively, this would indicate that such representations preserve important characteristics~from~the~initial~graph.

In the case of gravity-inspired graph AE, the encoder is a \textit{graph neural network} (GNN) \cite{hamilton2017representation} processing $A$ and $X$ (we discuss architecture choices in Section \ref{s341}), and returning an $n \times (d+1)$ matrix $\tilde{Z}$. The $i$-th row of $\tilde{Z}$ is a $(d+1)$-dimensional vector $\tilde{z}_i$. The $d$ first dimensions of $\tilde{z}_i$ correspond to the embedding vector $z_i$ of node $i$; the last dimension corresponds to the mass\footnote{Learning $\tilde{m}_{i}$, as defined in equation (\ref{equation2}), is equivalent to learning $m_i$, but allows to get rid of the logarithm and of the constant $G$ in computations.} $\tilde{m}_{i}$. Then, a decoder reconstructs $\hat{A}$ from the aforementioned acceleration formula:
\begin{equation}
\tilde{Z} = \text{GNN}(A,X), \text{ then }  \forall (i,j) \in \{1,...,n\}^2, \hat{A}_{ij} = \sigma(\tilde{m}_{j} - \log \|z_i - z_j\|_2^2).
\end{equation}
During training, the GNN's weights are tuned by minimizing a \textit{reconstruction loss} capturing the quality of the reconstruction $\hat{A}$ w.r.t. the true $A$, using gradient descent \cite{goodfellow2016deep}. In \cite{salha2019-2}, this loss is formulated as a standard \textit{weighted~cross~entropy} ~\cite{salha2020simple}.

\subsubsection{Gravity-Inspired Graph VAE} 
\label{s333}

Introduced as a probabilistic variant of gravity-inspired graph AE, \textit{gravity-inspired graph variational autoencoders (VAE)} \cite{salha2019-2} extend the graph VAE model from Kipf and Welling \cite{kipfvgae}, originally designed for undirected graphs. They provide an alternative strategy to learn $\tilde{Z}$, assuming that $\tilde{z}_i$ vectors are drawn from Gaussian distributions - one for each node - that must be learned. Formally, they  consider the following inference~model ~\cite{salha2019-2}, acting~as~an~encoder: $q(\tilde{Z}|A,X) = \prod_{i=1}^n q(\tilde{z}_i|A,X)$, with
$q(\tilde{z}_i|A,X) = \mathcal{N}(\tilde{z}_i|\mu_i, \text{diag}(\sigma_i^2))$.

Gaussian parameters are learned from two GNNs, i.e. $\mu = \text{GNN}_{\mu}(A,X)$, with $\mu$ denoting the $n \times (d+1)$ matrix stacking up mean vectors $\mu_i$. Similarly, $\log \sigma = \text{GNN}_{\sigma}(A,X)$.
Finally, a generative model decodes $A$ using, as above, the acceleration formula: 
 $p(A|\tilde{Z}) = \prod_{i=1}^n \prod_{j=1}^n p(A_{ij}|\tilde{z}_i, \tilde{z}_j)$,  with $\textbf{} p(A_{ij}|\tilde{z}_i, \tilde{z}_j) = \hat{A}_{ij} = \sigma(\tilde{m}_{j} - \log \|z_i - z_j\|_2^2)$.

During training, weights from the two GNN encoders are tuned by maximizing, as a reconstruction quality measure, a tractable \textit{variational lower bound (ELBO) of the model's likelihood} \cite{kipfvgae,salha2019-2}, using gradient descent. We refer to \cite{salha2019-2} for the exact formula of this ELBO loss, and to \cite{kingma2013vae,doersch2016tutorial} for details on derivations of variational bounds for VAE models. 

Besides constituting generative models with powerful applications to various graph generation problems \cite{molecule1,molecule2}, graph VAE models emerged as competitive alternatives to graph AE on some link prediction problems \cite{salha2019-2,salha2020simple,semiimplicit2019,kipfvgae}. We therefore saw value in considering both gravity-inspired graph AE and gravity-inspired graph VAE in our work.

\subsection{Cold Start Similar Items Ranking using Gravity-Inspired Graph AE/VAE}
\label{s34}

In this subsection, we now explain how we build upon these models to address cold start similar items ranking.

\subsubsection{Encoding Cold Nodes with GCNs}
\label{s341}

In this paper, our GNN encoders will be \textit{graph convolutional networks}~(GCN)~\cite{kipfgcn}. In a GCN with $L$ layers, with input layer $H^{(0)} = X$ corresponding to node attributes, and output layer $H^{(L)}$ (with $H^{(L)} = \tilde{Z}$ for AE, and $H^{(L)} = \mu$ or $\log \sigma$ for VAE), we have $H^{(l)} = \text{ReLU} (\tilde{A} H^{(l-1)} W^{(l-1)})$ for  $l \in \{1,...,L-1\}$ and $ H^{(L)} = \tilde{A} H^{(L-1)} W^{(L-1)}$. Here, $\tilde{A}$ denotes the \textit{out-degree normalized}\footnote{Formally, $\tilde{A} = D_{\text{out}}^{-1} (A + I_n)$ where $I_n$ is the $n \times n$ identity matrix and $D_{\text{out}}$ is the diagonal out-degree matrix \cite{salha2019-2} of $A + I_n$.} version of $A$ \cite{salha2019-2}. Therefore, at each layer, each node averages the representations from its neighbors (and itself), with a ReLU activation: $\text{ReLU}(x) = \max(x,0)$. $W^{(0)},...,W^{(L-1)}$ are the weight matrices to tune. More precisely, we will implement \textit{2-layer GCNs}, meaning~that,~for~AE:
\begin{equation}
\tilde{Z} = \tilde{A} \text{ReLU} (\tilde{A} X W^{(0)}) W^{(1)}.
\end{equation}
For VAE, $\mu  =\tilde{A} \text{ReLU} (\tilde{A} X W^{(0)}_{\mu}) W^{(1)}_{\mu}$, $\log \sigma  =\tilde{A} \text{ReLU} (\tilde{A} X W^{(0)}_{\sigma}) W^{(1)}_{\sigma}$, and $\tilde{Z}$ is then sampled from $\mu$ and $\log \sigma$.
As all outputs are $n \times (d+1)$ matrices and $X$ is an $n \times f$ matrix, then $W^{(0)}$ (or, $W^{(0)}_{\mu}$ and $W^{(0)}_{\sigma}$) is an $f \times d_{\text{hidden}}$ matrix, with $d_{\text{hidden}}$ the hidden layer dimension, and $W^{(1)}$ (or, $W^{(1)}_{\mu}$ and $W^{(1)}_{\sigma}$) is an $d_{\text{hidden}} \times (d+1)$~matrix.

We rely on GCN encoders as they permit \textit{incorporating new nodes}, attributed but isolated in the graph, into an existing embedding. Indeed, let us consider an autoencoder trained from some $A$ and $X$, leading to optimized weights $W^{(0)}$ and $W^{(1)}$ for some GCN. If $m \geq 1$ cold nodes appear, along with their $f$-dimensional description vectors:
\begin{itemize}
    \item $A$ becomes $A'$, an $(n+m) \times (n+m)$ adjacency matrix\footnote{Its $(n+m) \times (n+m)$ out-degree normalized version is $\tilde{A}' = D_{\text{out}}^{'-1} (A' + I_{(n+m)})$, where $D_{\text{out}}^{'-1}$ is the diagonal out-degree matrix of $A' + I_{(n+m)}$.}, with $m$ new rows and columns filled with zeros. 
    \item $X$ becomes $X'$, an $(n+m) \times f$ attribute matrix, concatenating $X$ and the $f$-dim descriptions of the $m$ new nodes.
    \item We derive embedding vectors and masses of new nodes through a \textit{forward pass}\footnote{We note that such GCN forward pass is possible since dimensions of weight matrices $W^{(0)}$ and $W^{(1)}$ are \textit{independent} of the number of nodes.} into the GCN previously trained on warm nodes, i.e. by computing the $(n+m) \times (d+1)$ new embedding matrix $\tilde{Z}' = \tilde{A}' \text{ReLU} (\tilde{A}' X' W^{(0)}) W^{(1)}$.
\end{itemize} 
We emphasize that the choice of GCN encoders is made without loss of generality. Our framework remains valid for any encoder similarly processing new attributed nodes. In our experiments, 2-layer GCNs reached better or comparable results w.r.t. some considered alternatives, namely deeper GCNs, linear encoders \cite{salha2020simple}, and graph~attention~networks~\cite{velivckovic2019graph}.

\subsubsection{Ranking Similar Items}
\label{s342}

After projecting cold nodes into the warm embedding, we use the \textit{gravity-inspired decoder} to predict their masked connections. More precisely, in our experiments, we add a hyperparameter $\lambda \in \mathbb{R}^+$ to equation~(\ref{equation2}) for flexibility. The \textit{estimated similarity weight} $\hat{A}_{ij}$ between some cold node $i$ and another node $j$ is thus:
\begin{equation}
\hat{A}_{ij} = \sigma(\underbrace{\tilde{m}_j}_{\text{influence of $j$}} - \lambda \times \underbrace{\log \|z_i - z_j\|_2^2}_{\text{proximity of $i$ and $j$}}).
\label{equation5}
\end{equation}
Then, the predicted top-$k$ most similar items of $i$ will correspond to the $k$ nodes $j$ with highest estimated weights $\hat{A}_{ij}$.

We interpret equation (\ref{equation5}) in terms of influence/proximity trade-off. The \textit{influence} part of (\ref{equation5}) indicates that, if two nodes $j$ and $l$ are equally close to $i$ in the embedding (i.e. $\|z_i - z_j\|_2 = \|z_i - z_l\|_2$), then $i$ will more likely points towards the node with the largest mass (i.e. \textit{"influence"}; we will compare these masses to popularity metrics in experiments). The \textit{proximity} part of (\ref{equation5}) indicates that, if $j$ and $l$ have the same mass (i.e. $\tilde{m}_j = \tilde{m}_l$), then $i$ will more likely points towards its closest neighbor, which could e.g. capture a closer musical similarity for artists. As illustrated in Section~\ref{s434}, tuning $\lambda$ will help us flexibly balance between these two aspects, and thus control for \textit{popularity biases} \cite{schedl2018current}~in~our~recommendations.

\subsubsection{On Complexity}
\label{s343}

Training models via full-batch gradient descent requires reconstructing the entire $\hat{A}$, which has an $O(dn^2)$ time complexity due to the evaluation of pairwise distances \cite{salha2019-1,salha2019-2}. While we will follow this strategy, our released code will also implement FastGAE \cite{salha2020fastgae}, a fast optimization method which approximates losses by decoding random subgraphs of $O(n)$ size.
Moreover, projecting cold nodes in an embedding only requires a single forward GCN pass, with linear time complexity w.r.t. the number of edges \cite{kipfgcn, salha2020simple}. This is another advantage of using GCNs w.r.t. more complex encoders. Last, retrieving the top-$k$ accelerations boils down to a nearest neighbors search in a $O(ndk)$ time, which could even be improved in future works with approximate search methods \cite{abbasifard2014survey}.

\section{Experimental Evaluation}
\label{s4}

We now present the experimental evaluation of our framework on music artists data from the Deezer production system.

\subsection{Experimental Setting: Ranking Similar Artists on a Music Streaming Service}
\label{s41}

\subsubsection{Dataset}  
\begin{figure*}[t]
  \centering
  \includegraphics[width=0.74\linewidth]{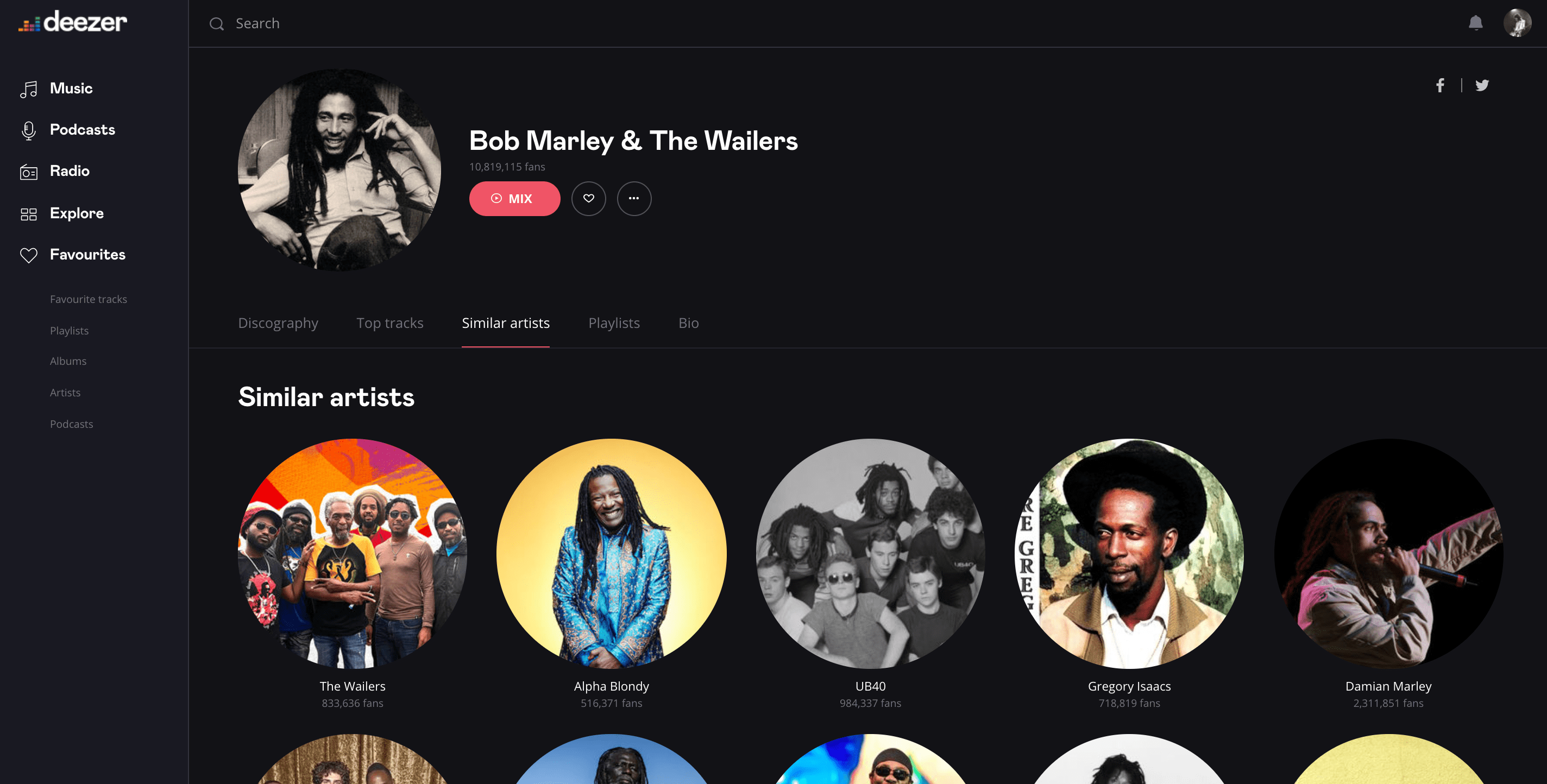}
  \caption{Similar artists recommended on the website version of Deezer. The mobile app proposes identical recommendations.}
  \label{deezerexample}
\end{figure*}

We consider a directed graph of 24 270 artists with various musical characteristics (see below), extracted from the music streaming service Deezer. Each artist points towards $k=20$ other artists. They correspond, up to internal business rules, to the top-20 artists from the same graph that would be recommended by our production system on top of the \textit{"Fans Also Like/Similar Artists"} feature illustrated in Figure~\ref{deezerexample}.
Each directed edge $(i,j)$ has a weight $A_{ij}$ normalized in the $[0,1]$~set; for unconnected pairs, $A_{ij}= 0$.  It corresponds to the similarity score of artist $j$ w.r.t. $i$, computed on a weekly basis from usage data of millions of Deezer users. More precisely, weights are based on \textit{mutual information} scores \cite{shakibian2017mutual} from \textit{artist co-occurrences among streams}. Roughly, they compare the probability that a user listens to the two artists, to their global listening frequencies on the service, and they are normalized at the artist level through internal heuristics and business rules (some details on exact score computations are voluntarily omitted for confidentiality reasons). In the graph, edges correspond to the 20 highest scores for each node. In general, $A_{ij} \neq A_{ji}$. In particular, $j$ might be the most similar artist of $i$ while $i$ does not appear among the top-20 of $j$.

We also have access to descriptions of these artists, either extracted through the musical content or provided by record labels. Here, each artist $i$  will be described by an attribute vector $x_i$ of dimension $f=56$, concatenating:
\begin{itemize}
    \item A 32-dimensional \textit{genre} vector. Deezer artists are described by music genres \cite{epure-etal-2020-modeling}, among more than~300. 32-dim embeddings are learned from these genres, by factorizing a co-occurrence matrix based on listening usages with SVD \cite{koren2009matrix}. Then, the genre vector of an artist is the average of embedding vectors of~his/her~music~genres.
    \item A 20-dimensional \textit{country} vector. It corresponds to a one-hot encoding vector, indicating the country of origin of an artist, among the 19 most common countries on Deezer, and with a 20\up{th} category gathering all other countries.
    \item A 4-dimensional \textit{mood} vector. It indicates the average and standard deviations of the \textit{valence} and \textit{arousal} scores across an artist's discography. In a nutshell, valence captures whether each song has a positive or negative mood, while arousal captures whether each song has a calm or energetic mood \cite{russell1980circumplex,delbouys2018music,huang2016bi}. These scores are computed internally, from audio data and using a deep neural network inspired from the work of Delbouys et al. \cite{delbouys2018music}.
\end{itemize}
While some of these features are quite general, we emphasize that the actual Deezer app also gathers more refined information on artists, e.g. from audio or textual descriptions. They are undisclosed and unused in these~experiments.

\subsubsection{Problem} 
\label{s412}
We consider the following similar artists ranking problem.
Artists are split into a training set, a validation set and a test set gathering 80\%, 10\% and 10\% of artists respectively. The training set corresponds to warm artists. Artists from the validation and test sets are the cold nodes: their edges are \textit{masked}, and they are therefore observed as isolated in the graph. Their 56-dimensional descriptions are available. We evaluate the ability of our models at retrieving these edges, with correct weight ordering. As a measure of \textit{prediction accuracy}, we will report \textit{Recall@K} scores. They indicate, for various $K$, which proportion of the 20 ground-truth similar artists appear among the top-$K$ artists with largest estimated weights. Moreover, as a measure of \textit{ranking quality}, we will also report the widely used \textit{Mean Average Precision~at}~$K$~(\textit{MAP@K}) and \textit{Normalized Discounted Cumulative Gain
}~at~$K$~(\textit{NDCG@K}) scores\footnote{MAP@K and NDCG@K are computed as in  equation (4) of \cite{schedl2018current} (averaged over all cold artists) and in equation (2) of \cite{wang2013theoretical} respectively.}.

\subsubsection{Dataset and Code Release} We publicly release our industrial data (i.e. the complete graph, all weights and all descriptive vectors) as well as our source code on GitHub\footnote{Code and data will be available by the RecSys 2021 conference on: \href{https://github.com/deezer/similar_artists_ranking}{https://github.com/deezer/similar\_artists\_ranking}}. Besides making our results fully reproducible, such a release publicly provides a new benchmark dataset to the research community, permitting the evaluation of comparable graph-based recommender systems on~real-world~resources.

\subsection{List of Models and Baselines}
\label{s42}
We now describe all methods considered in our experiments. All embeddings have $d = 32$, which we will discuss. Also, all hyperparameters mentioned thereafter were tuned by optimizing NDCG@20 scores on the validation set.

\subsubsection{Gravity-Inspired Graph AE/VAE} We follow our framework from Section \ref{s34} to embed cold nodes. For both AE and VAE, we use 2-layer GCN encoders with 64-dim hidden layer, and 33-dim output layer (i.e. 32-dim $z_i$ vectors, plus the mass), trained for 300 epochs. We use Adam optimizer \cite{kingma2014adam}, with a learning rate of 0.05, without dropout, performing full-batch gradient descent, and using the reparameterization trick \cite{kingma2013vae} for VAE. We set $\lambda$ = 5  in the decoder of~(\ref{equation5}) and discuss the impact of $\lambda$ thereafter. Our adaptation of these models builds upon the Tensorflow~code~of~Salha~et~~al.~~\cite{salha2019-2}.

\subsubsection{Other Methods based on the Directed Artist Graph} We compare gravity-inspired graph AE/VAE to \textit{standard graph AE /VAE} models~\cite{kipfgcn}, with a similar setting as above. These models use symmetric inner-product decoders i.e. $\hat{A}_{ij} = \sigma(z^T_i z_j)$, therefore ignoring directionalities. Moreover, we implement \textit{source-target graph AE/VAE}, used as a baseline in \cite{salha2019-2}. They are similar to standard graph AE/VAE, except that they decompose the 32-dim vectors $z_i$ into a \textit{source} vector $z^{(s)}_i = z_{i[1: 16]}$ and a \textit{target} vector $z^{(t)}_i = z_{i[17:32]}$, and then decode edges as follows: $\hat{A}_{ij} = \sigma(z^{(s)T}_i z^{(t)}_j)$ and $\hat{A}_{ji} = \sigma(z^{(s)T}_j z^{(t)}_i)$ ($\neq \hat{A}_{ij}$ in general). They reconstruct directed links, as gravity models, and are therefore relevant baselines for our evaluation. Last, we also test the recent DEAL model \cite{ijcai2020DEAL} mentioned in Section \ref{s22}, and designed for inductive link prediction on new isolated but attributed nodes. We used the authors' PyTorch implementation, with similar attribute and structure encoders, alignment mechanism, loss and cosine predictions as their original~model~\cite{ijcai2020DEAL}.

\subsubsection{Other Baselines} 

In addition, we compare our proposed framework to four popularity-based baselines:
\begin{itemize}
    \item \textit{Popularity}: recommends the $K$ most popular\footnote{Our dataset includes the \textit{popularity} rank (from 1\up{st} to $n$\up{th}) of warm artists. It is out of $x_i$ vectors, as it is usage-based and thus unavailable for cold~artists.} artists on Deezer (with $K$ as~in~Section~\ref{s412}).
    \item \textit{Popularity by country}: recommends the $K$ most popular artists \textit{from the country of origin} of the cold artist.
    \item \textit{In-degree}: recommends the $K$ artists with highest \textit{in-degrees} in the graph, i.e. sum of weights pointing to them. 
    \item \textit{In-degree by country}: proceeds as \textit{In-degree}, but on warm artists from the country of origin of the cold artist.
\end{itemize}
We also consider three baselines only or mainly based on musical descriptions $x_i$ and not on usage data:
\begin{itemize}
    \item \textit{$K$-NN}: recommends the $K$ artists with closest $x_i$ vectors, from a nearest neighbors search with Euclidean distance.
    \item \textit{$K$-NN + Popularity} and \textit{$K$-NN + In-degree}: retrieve the 200 artists with closest $x_i$ vectors, then recommends the $K$ most popular ones among these 200 artists, ranked according to \textit{popularity} and \textit{in-degree} values respectively.
\end{itemize}
We also implement \textit{SVD+DNN}, which follows the \textit{"embedding+mapping"} strategy from Section~\ref{s22} by 1) computing an SVD \cite{koren2009matrix} of the warm artists similarity matrix, learning 32-dim $z_i$ vectors, 2) training a 3-layer neural network (with layers of dimension 64, 32 and 32, trained with Adam \cite{kingma2014adam} and a learning rate of 0.01) to map warm $x_i$ vectors to $z_i$ vectors, and 3) projecting cold artists into the SVD embedding through this mapping. Last, among deep learning approaches from Section \ref{s22} (CVAE, DropoutNet, MeLU, STAR-GCN), we report results from the two best methods on our dataset, namely \textit{DropoutNet} \cite{volkovs2017dropoutnet} and \textit{STAR-GCN} \cite{zhang2019star}, using the authors' implementations with careful fine-tuning on validation artists\footnote{These last models do not process similar artist graphs, but raw \textit{user-item usage data}, either as a bipartite user-artist graph or as an interaction matrix. While Deezer can not release such fine-grained data, we nonetheless provide embedding vectors from these baselines to reproduce our scores.}. Similar artists ranking is done via a nearest neighbors search in the~resulting~embedding~spaces.

\subsection{Results}
\label{s43}

\begin{table*}[!t]
\centering
\caption{Cold start similar artists ranking: performances on test set.}
\resizebox{1.0\textwidth}{!}{
\begin{tabular}{r|ccc|ccc|ccc}
\toprule
\textbf{Methods} & \multicolumn{3}{c}{\textbf{Recall@K (in \%)}} & \multicolumn{3}{c}{\textbf{MAP@K (in \%)}} & \multicolumn{3}{c}{\textbf{NDCG@K (in \%)}} \\
($d = 32$)&  \scriptsize \textbf{$K=20$} & \scriptsize \textbf{$K=100$} & \scriptsize \textbf{$K=200$} & \scriptsize \textbf{$K=20$} & \scriptsize \textbf{$K=100$} & \scriptsize \textbf{$K=200$} & \scriptsize \textbf{$K=20$} & \scriptsize \textbf{$K=100$} & \scriptsize \textbf{$K=200$} \\
\midrule
\midrule 
Popularity & 0.02 & 0.44 & 1.38 & <0.01 & 0.03 & 0.12 & 0.01 & 0.17 & 0.44 \\ 
Popularity by country & 2.76 & 12.38 & 18.98 & 0.80 & 3.58 & 6.14 & 2.14 & 6.41 & 8.76 \\ 
In-degree & 0.91 & 3.43 & 6.85 & 0.15 & 0.39 & 0.86 & 0.67 & 1.69 & 2.80 \\ 
In-degree by country & 5.46 & 16.82 & 23.52 & 2.09 & 5.43 & 7.73 & 5.00 & 10.19 & 12.64 \\ 
$K$-NN on $x_i$ & 4.41 & 13.54 & 19.80 & 1.14 & 3.38 & 5.39 & 4.29 & 8.83 & 11.22\\ 
$K$-NN + Popularity & 5.73 & 15.87 & 19.83 & 1.66 & 4.32 & 5.74 & 4.86 & 10.03 & 11.76 \\ 
$K$-NN + In-degree & 7.49 & 17.29 & 18.76 & 2.78 & 5.60 & 6.18 & 7.41 & 12.48 & 13.14  \\ 
SVD + DNN  & 6.42 $\pm$ 0.96 & 21.83 $\pm$ 1.21 & 35.01 $\pm$ 1.41 & 2.25 $\pm$ 0.67 & 6.36 $\pm$ 1.19 & 11.52 $\pm$ 1.98 & 6.05 $\pm$ 0.75 & 12.91 $\pm$ 0.92 & 17.89 $\pm$ 0.95 \\ 
STAR-GCN & 10.03 $\pm$ 0.56 & 31.45 $\pm$ 1.09 & 43.92 $\pm$ 1.10 & 3.10 $\pm$ 0.32 & 10.64 $\pm$ 0.54 & 16.62 $\pm$ 0.68 & 10.07 $\pm$ 0.40 & 21.17 $\pm$ 0.69 & 25.99 $\pm$ 0.75 \\ 
DropoutNet & 12.96 $\pm$ 0.54 & 37.59 $\pm$ 0.76 & 49.93 $\pm$ 0.82 & 4.18 $\pm$ 0.30 & 13.61 $\pm$ 0.55 & 20.12 $\pm$ 0.67 & 13.12 $\pm$ 0.68 & 25.61 $\pm$ 0.72 & 30.52 $\pm$ 0.78 \\ 
DEAL  & 12.80 $\pm$ 0.52 & 37.98 $\pm$ 0.59 & 50.75 $\pm$ 0.72 & 4.15 $\pm$ 0.25 & 14.01 $\pm$ 0.44 & 20.92 $\pm$ 0.54 & 12.78 $\pm$ 0.53 & 25.70 $\pm$ 0.62 & 30.69 $\pm$ 0.70 \\ 
Graph AE & 7.30 $\pm$ 0.51 & 25.92 $\pm$ 0.95 & 40.37 $\pm$ 1.11 & 2.81 $\pm$ 0.29 & 7.97 $\pm$ 0.47 & 14.24 $\pm$ 0.67 & 6.32 $\pm$ 0.39 & 15.54 $\pm$ 0.66 & 20.94 $\pm$ 0.72 \\ 
Graph VAE & 10.01 $\pm$ 0.52 & 34.00 $\pm$ 1.06 & 49.72 $\pm$ 1.14 & 3.53 $\pm$ 0.27 & 11.68 $\pm$ 0.52 & 19.46 $\pm$ 0.70 & 10.09 $\pm$ 0.58 & 21.37 $\pm$ 0.73 & 27.31 $\pm$ 0.75 \\ 
Sour.-Targ. Graph AE & 12.21 $\pm$ 1.30 & 39.52 $\pm$ 3.53 & 56.25 $\pm$ 3.57 & 4.62 $\pm$ 0.81 & 14.67 $\pm$ 2.33 & 23.60 $\pm$ 2.85 & 12.42 $\pm$ 1.39 & 25.45 $\pm$ 3.37 & 31.80 $\pm$ 3.38 \\ 
Sour.-Targ. Graph VAE & 13.52 $\pm$ 0.64 & 42.68 $\pm$ 0.69 & 59.51 $\pm$ 0.76 & 5.19 $\pm$ 0.31 & 16.07 $\pm$ 0.40 & 25.48 $\pm$ 0.55 & 13.60 $\pm$ 0.73 & 27.81 $\pm$ 0.56 & 34.19 $\pm$ 0.59 \\ 
\midrule%
\textbf{Gravity Graph AE} & \textbf{18.33} $\pm$ \textbf{0.45} & \textbf{52.26} $\pm$ \textbf{0.90} & \textbf{67.85} $\pm$ \textbf{0.98} & \textbf{6.64} $\pm$ \textbf{0.25} & \textbf{21.19} $\pm$ \textbf{0.55} & \textbf{30.67} $\pm$ \textbf{0.68} & \textbf{18.64} $\pm$ \textbf{0.47} & \textbf{35.77} $\pm$ \textbf{0.66} & \textbf{41.42} $\pm$ \textbf{0.68} \\ 
\textbf{Gravity Graph VAE} & 16.59 $\pm$ 0.50 & 49.51 $\pm$ 0.78 & 65.70 $\pm$ 0.75 & 5.66 $\pm$ 0.35 & 19.07 $\pm$ 0.57 & 28.66 $\pm$ 0.59 & 16.74 $\pm$ 0.55 & 33.34 $\pm$ 0.66 & 39.29 $\pm$ 0.64 \\ 
\bottomrule
\end{tabular}}
\label{tableresults}
\end{table*}

\subsubsection{Performances} Table~\ref{tableresults} reports mean performance scores for all models, along with standard deviations
over 20 runs for models with randomness due to weights initialization in GCNs or neural networks. Popularity and In-degree appear as the worst baselines. Their scores significantly improve by focusing on the country of origin of cold artists (e.g. with a Recall@100 of 12.38\% for Popularity by country, v.s. 0.44\% for Popularity). Besides, we observe that methods based on a direct $K$-NN search from $x_i$ attributes are outperformed by the more elaborated cold start methods leveraging both attributes and warm usage data. In particular, DropoutNet, as well as the graph-based DEAL, reach stronger results than SVD+DNN and STAR-GCN. They also surpass standard graph AE and VAE (e.g. with a +6.46 gain in average NDCG@20 score for DEAL w.r.t. graph AE), but not the AE/VAE extensions that explicitly model edges directionalities, i.e. source-target graph AE/VAE and, even more, gravity-inspired graph AE/VAE, that provide the best recommendations. It emphasizes the effectiveness of our framework, both in terms of prediction accuracy (e.g. with a top 67.85\% average Recall@200 for gravity-inspired graph AE) and of ranking quality (e.g. with a top 41.42\% average NDCG@200 for this same method). Moreover, while embeddings from Table~\ref{tableresults} have $d =32$, we point out that gravity-inspired models remained superior on our tests with $d=64$ and $d=128$. Also, VAE methods tend to outperform their deterministic counterparts for standard and source-target models, while the contrary conclusion emerges on gravity-inspired models. This confirms the value of considering both settings when testing these models on novel applications. 

\subsubsection{On the mass parameter} 
\begin{figure*}[t]
\minipage{0.58\textwidth}
\centering
  \includegraphics[width=0.9\linewidth]{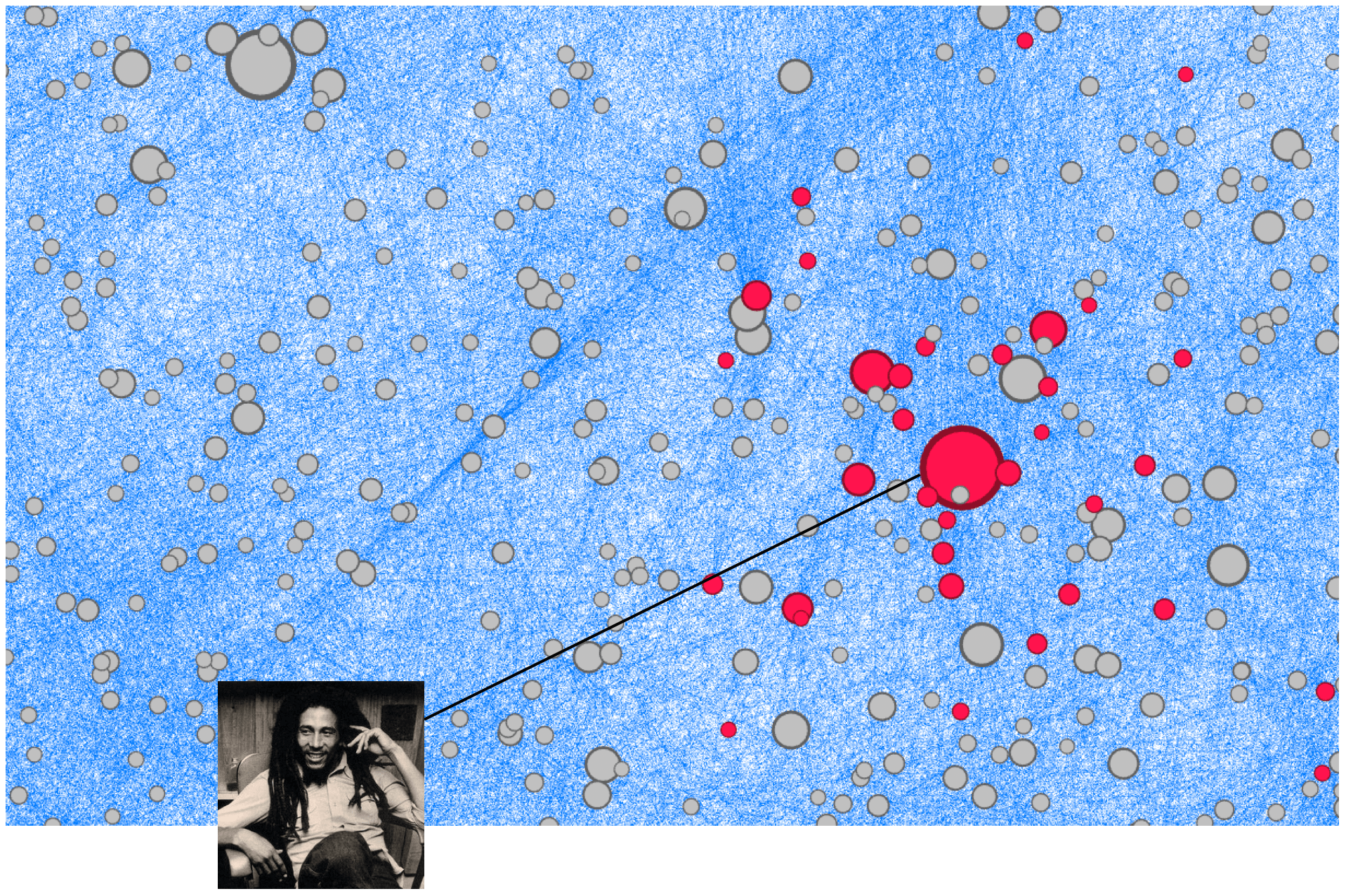}
  \caption{Visualization of some music artists from gravity-inspired graph AE. Nodes are scaled using masses $\tilde{m}_i$, and node separation is based on distances in the embedding, using multidimensional scaling and with \cite{fruchterman1991graph}. Red nodes correspond to \textit{Jamaican reggae} artists, appearing in the same neighborhood.}
  \label{visualization}
\endminipage\hfill
\minipage{0.4\textwidth}
\centering
  \centering
  \includegraphics[width=0.91\linewidth]{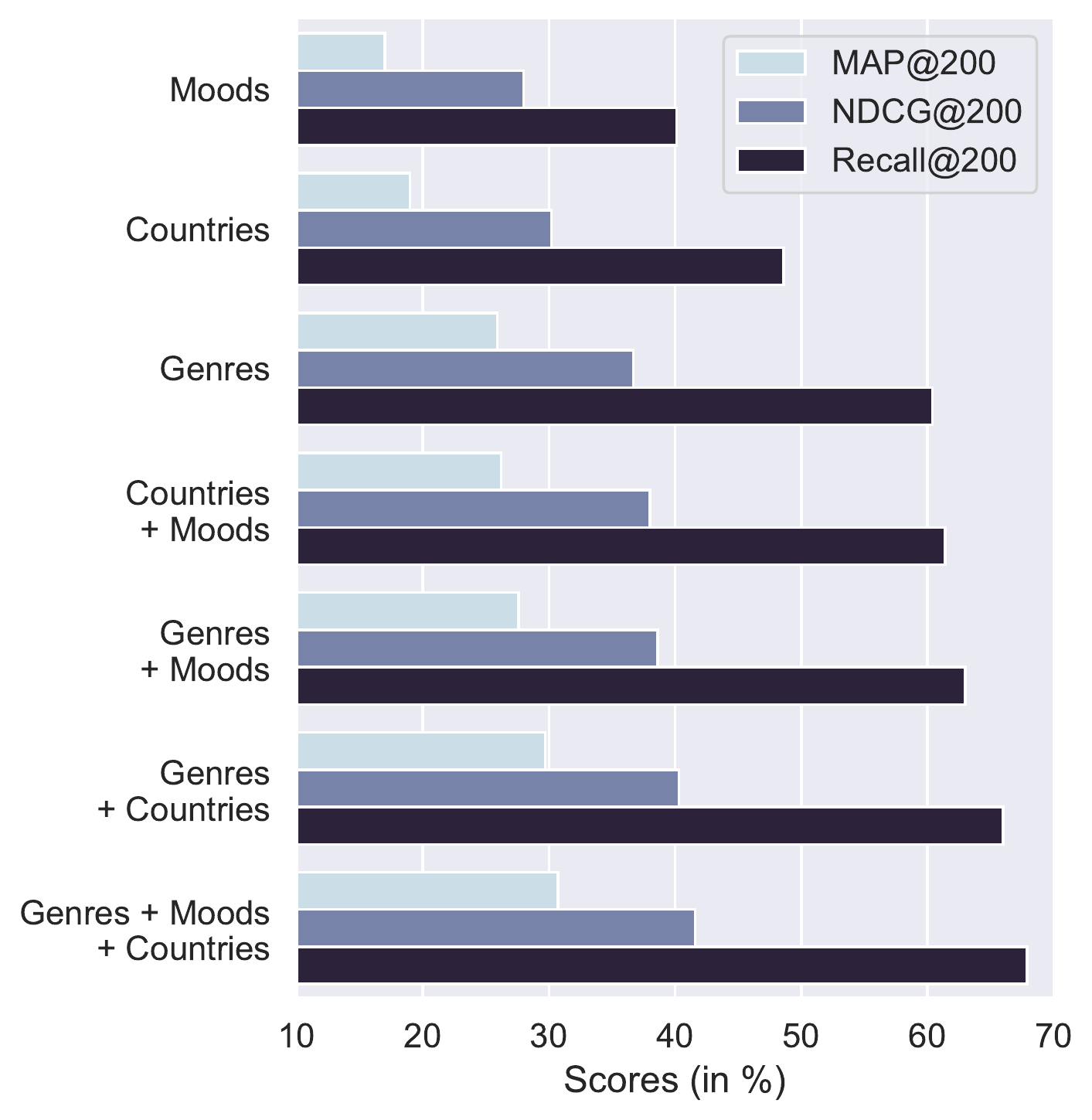}
  \caption{Performances of gravity-inspired graph AE, when only trained from subsets of artist-level attributes, among countries, music genres and moods.}
  \label{impactfeatures}
\endminipage
\end{figure*}

In Figure~\ref{visualization}, we visualize some artists and their estimated masses. At first glance, one might wonder whether nodes $i$ with largest masses~$\tilde{m}_i$, as Bob~Marley in Figure~\ref{visualization}, simply correspond to the most popular artists on Deezer. Table~\ref{correlations} shows that, while masses are indeed positively correlated to popularity and to various graph-based node importance measures, these correlations are not perfect, which highlights that our models do not exactly learn any of these metrics. 
Furthermore, replacing masses $\tilde{m}_i$  by any of these measures during training (i.e. by optimizing $z_i$ vectors with the mass of $i$ being fixed, e.g. as its PageRank \cite{page1999pagerank} score) diminishes performances (e.g. more than -6 points in NDCG@200, in the case of PageRank), which confirms that jointly learning embeddings and masses is optimal. Last, qualitative investigations also tend to reveal that masses, and thus relative node attractions, vary across location in the graph. Local influence correlates with popularity but is also impacted by various culture or music-related factors such as countries and genres. As an illustration, the successful samba/pagode Brazilian artist Thiaguinho, out of the top-100 most popular artists from our training set, has a larger mass than American pop star Ariana Grande, appearing among the top-5 most popular ones. While numerous Brazilian pagode artists point towards Thiaguinho, American pop music is much broader and all pop artists do not point towards Ariana Grande despite her popularity.

\subsubsection{Impact of attributes} So far, all models processed the complete 56-dimensional attribute vectors $x_i$, concatenating information on music genres, countries and moods. In Figure~\ref{impactfeatures}, we assess the actual impact of each of these descriptions on performances, for our gravity-inspired graph VAE. Assuming only one attribute (genres, countries \textit{or} moods) is available during training, genres-aware models return the best performances. Moreover, adding moods to country vectors leads to larger gains than adding moods to genre vectors. This could reflect how some of our music genres, such as \textit{speed metal}, already capture some valence or arousal characteristics. Last, Figure~\ref{impactfeatures} confirms that gathering all three descriptions provides the best performances, corresponding to those reported in Table~\ref{tableresults}. 

\subsubsection{Popularity/diversity trade-off} \label{s434}
\begin{figure*}[t]

\minipage{0.48\textwidth}
\centering
\begin{footnotesize}
\captionof{table}{Pearson and Spearman correlation coefficients of masses~$\tilde{m}_i$ from gravity-inspired graph AE, w.r.t. artist-level reversed popularity ranks (i.e. the higher the more popular) on Deezer and to three node importance measures: in-degree (i.e. sum of edges coming into the node), betweenness centrality \cite{salha2019-2} and PageRank \cite{page1999pagerank}. Coefficients computed on the~training~set.}
\label{correlations}
\begin{tabular}{c|c|c}
    \toprule
\textbf{Node-level} & \textbf{Pearson} & \textbf{Spearman} \\ \textbf{measures} & \textbf{correlation} & \textbf{correlation}\\
\midrule
\midrule
\textbf{Popularity Rank} & $0.208$ & $0.206$ \\
\textbf{Popularity Rank by Country} & $0.290$ & $0.336$ \\
\midrule
\textbf{In-degree Centrality} & $0.201$ & $0.118$ \\
\textbf{Betweenness Centrality} & $0.109$ & $0.079$\\
\textbf{PageRank Score} & $0.272$ & $0.153$\\
    \bottomrule
\end{tabular}
\end{footnotesize}
\endminipage\hfill
\minipage{0.51\textwidth}
\centering
  \centering
  \includegraphics[width=0.78\linewidth]{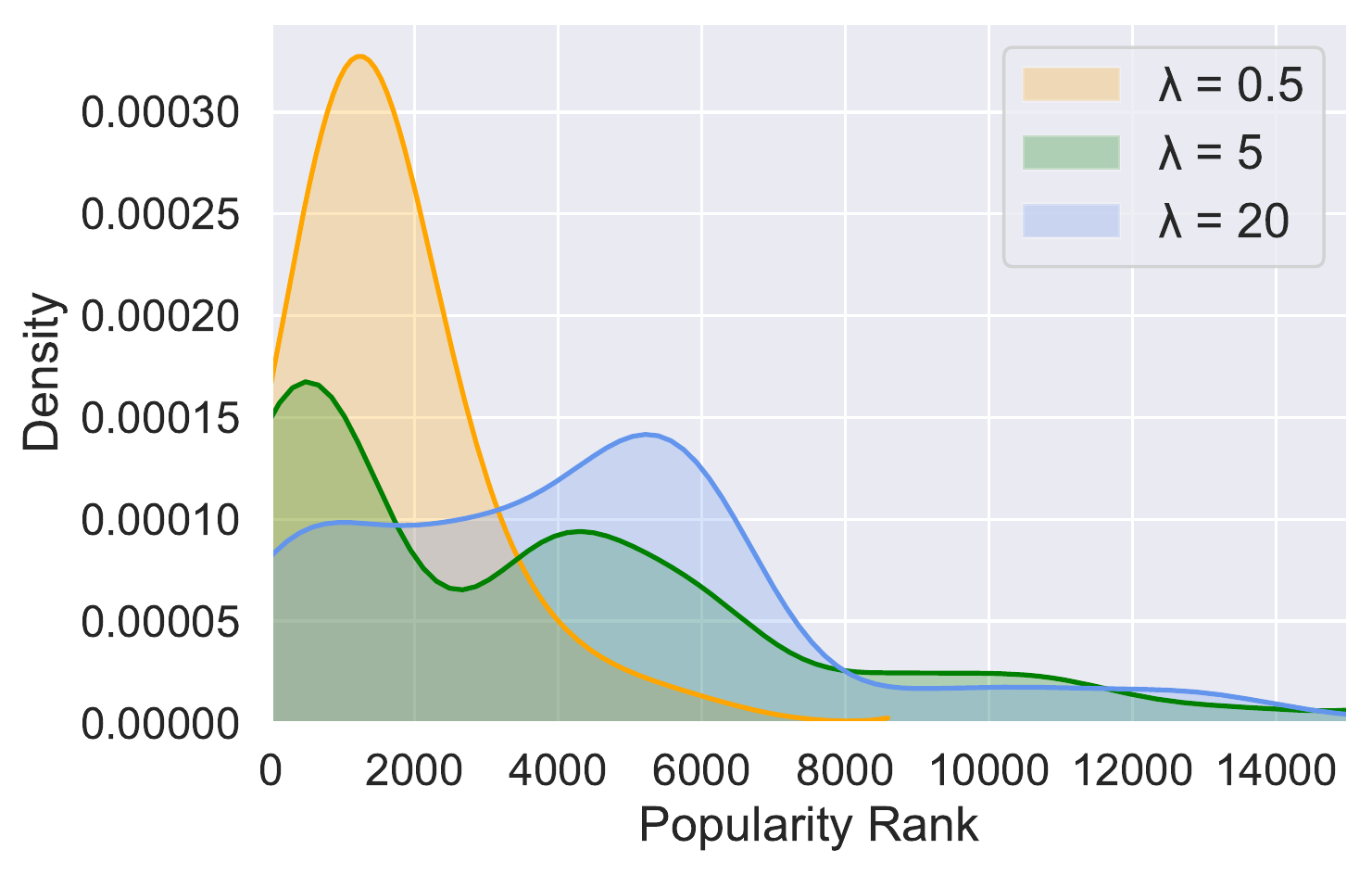}
  \caption{Popularity rank of the \textit{most popular} artist recommended to each test artist, among their top-20. Distributions obtained from gravity-inspired graph AE models with varying hyperparameters $\lambda$.}
  \label{impactlambda}
\endminipage
\end{figure*}

Last, besides performances, the gravity-inspired decoder from equation (\ref{equation5}) also enables us to flexibly address popularity biases when ranking similar artists. More precisely:
\begin{itemize}
    \item Setting $\lambda \rightarrow 0$ increases the relative importance of the \textit{influence} part of equation (\ref{equation5}). Thus, the model will highly rank the most massive nodes. As illustrated in Figure~\ref{impactlambda}, this results in recommending \textit{more popular} music artists.
    \item On the contrary, increasing $\lambda$ diminishes the relative importance of masses in predictions, in favor of the actual node \textit{proximity}. As illustrated in Figure~\ref{impactlambda}, this tends to increase the recommendation of~\textit{less~popular}~content.
\end{itemize}
Setting $\lambda = 5$ leads to optimal scores in our application (e.g. with a 41.42\% NDCG@200 for our AE, v.s. 35.91\% and 40.31\% for the same model with $\lambda=1$ and $\lambda=20$ respectively). Balancing between popularity and diversity is often desirable for industrial-level recommender systems \cite{schedl2018current}. Gravity-inspired decoders flexibly permit such a balancing.

\subsubsection{Possible improvements (on models)} Despite promising results, assuming fixed similarity scores over time might sometimes be unrealistic, as some user preferences could actually evolve. Capturing such changes, e.g. through \textit{dynamic graph embeddings}, might permit providing even more refined recommendations. Also, during training we kept $k=20$ edges for each artist, while one could consider varying $k$ at the artist level, i.e. adding more (or fewer) edges, depending on the actual musical relevance of each link. One could also compare different metrics, besides mutual information, when constructing the ground-truth graph. Last, we currently rely on a single GCN forward pass to embed cold artists which, while being fast and simple, might also be limiting. Future studies on more elaborated approaches, e.g. to incrementally update GCN weights when new nodes appear, could also improve our current framework.

\subsubsection{Possible improvements (on evaluation)} Our evaluation focused on the prediction of \textit{ranked lists for cold artists}. This permits filling up their \textit{"Fans Also Like/Similar Artists"} sections, which was our main goal in this paper. On the other hand, future internal investigations could also aim at measuring to which extent the inclusion of new nodes in the embedding space impacts the existing \textit{ranked lists for warm artists}. Such an additional evaluation, e.g. via an online A/B test on Deezer, could assess which cold artists actually enter these lists, and whether the new recommendations 1)~are more diverse, according to some music or culture-based criteria, and 2) improve user engagement on the service.

\section{Conclusion}
\label{s5}

In this paper, we modeled the challenging cold start similar items ranking problem as a link prediction task, in a directed and attributed graph summarizing information from \textit{"Fans Also Like/Similar Artists"} features. We presented an effective framework to address this task, transposing recent advances on gravity-inspired graph autoencoders to recommender systems. Backed by in-depth experiments on artists from the global music streaming service Deezer, we emphasized the practical benefits of our approach, both in terms of recommendation accuracy, of ranking quality and of flexibility. Along with this paper, we publicly release our source code, as well as Deezer data from our experiments. We hope that this release of industrial resources will benefit future research on graph-based cold start recommendation. In particular, we already identified several directions that, in future works, should lead towards the improvement of our approach.

\bibliographystyle{ACM-Reference-Format}
\bibliography{references}

\end{document}